\documentclass[12pt]{article}
\pdfoutput=1
\usepackage{amsmath, amsfonts, amssymb, amsthm, bm}
\usepackage{lscape}
\usepackage[authoryear]{natbib}
\usepackage{algorithm, algpseudocode}
\usepackage{enumitem}
\usepackage{url}
\usepackage{graphicx}
\usepackage[labelformat = simple]{subcaption}

\newcommand{\lr}[3]{\left#1#3\right#2}
\newcommand{\norm}[1]{\Vert#1\Vert}
\newcommand{\abs}[1]{\vert#1\vert}
\DeclareMathOperator{\diag}{diag}

\renewcommand{\Re}{\mathbb{R}}

\newcommand{\dx}{\text{d}\boldsymbol{x}}

\newcommand{\captionfonts}{\normalsize}

\makeatletter  
\long\def\@makecaption#1#2{%
  \vskip\abovecaptionskip
  \sbox\@tempboxa{{\captionfonts #1: #2}}%
  \ifdim \wd\@tempboxa >\hsize
    {\captionfonts #1: #2\par}
  \else
    \hbox to\hsize{\hfil\box\@tempboxa\hfil}%
  \fi
  \vskip\belowcaptionskip}
\makeatother   
\begin{document}
\hspace{13.9cm}1

\ \vspace{20mm}\\

{\LARGE Regularized Multi-Task Learning for Multi-Dimensional Log-Density Gradient Estimation}

\ \\
{\bf \large
  Ikko Yamane,\ 
  Hiroaki Sasaki,\ 
  Masashi Sugiyama\\
}
{Graduate School of Frontier Sciences, The University of Tokyo, Japan.}\\
%

%\ \\[-2mm]

\thispagestyle{empty}
\markboth{}{NC instructions}
\ \vspace{-0mm}\\
\sloppy

\begin{center} {\bf Abstract} \end{center}
\emph{Log-density gradient estimation} is a fundamental statistical problem and possesses various practical applications such as clustering and measuring non-Gaussianity.
A naive two-step approach of first estimating the density and then taking its log-gradient is unreliable because an accurate density estimate does not necessarily lead to an accurate log-density gradient estimate.
To cope with this problem, a method to \emph{directly} estimate the log-density gradient without density estimation has been explored,
and demonstrated to work much better than the two-step method.
The objective of this paper is to further improve the performance of this direct method in multi-dimensional cases.
Our idea is to regard the problem of log-density gradient estimation in each dimension as a task,
and apply \emph{regularized multi-task learning} to the direct log-density gradient estimator.
We experimentally demonstrate the usefulness of the proposed multi-task method in log-density gradient estimation and mode-seeking clustering.

%1-----------------------------------------
\section{Introduction}\label{sec:intro}
\emph{Multi-task learning} is a paradigm of machine learning for solving multiple related learning tasks simultaneously
with the expectation that information brought by other related tasks can be mutually exploited to improve the accuracy~\citep{Caruana97}.
Multi-task learning is particularly useful when one has many related learning tasks to solve but only few training samples are available for each task, which is often the case in many real-world problems
such as therapy screening~\citep{bickel2008multi} and face verification~\citep{wang2009boosted}.

Multi-task learning has been gathering a great deal of attention, and extensive studies have been conducted both theoretically and experimentally~\citep{Thrun,RMTL,Ando05,Zhang13,BaxterInductiveBias}.
% Existing research
\citet{Thrun} proposed the \emph{lifelong learning framework}, which transfers the knowledge obtained from the tasks experienced in the past to a newly given task, and it was demonstrated to improve the performance of image recognition.
Baxter~\citet{BaxterInductiveBias} defined a multi-task learning framework called \emph{inductive bias learning},
and derived a generalization error bound.
The semi-supervised multi-task learning method proposed by \citet{Ando05} generates many auxiliary learning tasks from unlabeled data and seeks a good feature mapping for the target learning task.
%----- Regularized multi-task learning -----%
Among various methods of multi-task learning, one of the simplest and
most practical approaches would be \emph{regularized multi-task
learning}~\citep{RMTL,evgeniou2005learning}, which uses a regularizer that imposes the solutions of related tasks to be close to each other.
Thanks to its generic and simple formulation, regularized multi-task learning has been applied to various types of learning problems such as regression and classification~\citep{RMTL,evgeniou2005learning}.
In this paper, we explore a novel application of regularized multi-task learning to the problem of \emph{log-density gradient estimation}~\citep{Beran76,Cox85,Sasaki14}.

The goal of log-density gradient estimation is to estimate the gradient of the logarithm of an unknown probability density function using samples following it.
Log-density gradient estimation has various applications such as clustering~\citep{fukunaga1975mean,cheng1995mean,comaniciu2002mean,Sasaki14}, measuring non-Gaussianity~\citep{huber1985projection}
%non-parametric density estimation \citep{silverman1986density},
%divergence estimation \citep{noh2014bias},
and other fundamental statistical topics~\citep{singh1977applications}.

\citet{Beran76} proposed a method for \emph{directly} estimating gradients without going through density estimation, to which we refer as \emph{least-squares log-density gradients} (LSLDG).
This direct method was experimentally shown to outperform the naive one
consisting of density estimation followed by log-gradient computation,
and was demonstrated to be useful in clustering~\citep{Sasaki14}.

The objective of this paper is to estimate log-density gradients further accurately
in multi-dimensional cases, which is still a challenging topic even using LSLDG.
It is important to note that
since the output dimensionality of the log-density gradient $\nabla\log p(\bm x)$
is the same as its input dimensionality $d$,
multi-dimensional log-density gradient estimation can be regarded as 
having multiple learning tasks if we regard estimation of each output dimension as a task.
Based on this view, in this paper, we propose to apply regularized multi-task learning to LSLDG.
We also provide a practically useful design of parametric models for successfully applying regularized multi-task learning to log-density gradient estimation.
We experimentally demonstrate that the accuracy of LSLDG can be significantly improved by the proposed multi-task method in multi-dimensional log-density estimation problems and that a mode-seeking clustering method based on the proposed method outperforms other methods.

The organization of this paper is as follows: In
Section~\ref{sec:logDensityGradient}, we formulate the problem of
log-density gradient estimation and review LSLDG. Section~\ref{sec:RMTL}
reviews the core idea of regularized multi-task learning.
Section~\ref{sec:proposed} presents our proposed log-density gradient estimator and algorithms for computing the solution.
In Section~\ref{sec:gradExperiments}, we
experimentally demonstrate that the proposed method performs well on
both artificial and benchmark data. Application to mode-seeking
clustering is given in Section~\ref{sec:clustering}.
Section~\ref{sec:conclusion} concludes this paper with potential
extensions of this work.

%2-----------------------------------------
\section{Log-density gradient estimation}
\label{sec:logDensityGradient}
In this section, we formulate the problem of \emph{log-density gradient
estimation}, and then review LSLDG.
\subsection{Problem formulation and a naive method}
\begin{figure*}[t]
  \centering
  \begin{subfigure}{0.45\textwidth}
    \centering
    \includegraphics[width=\hsize,clip]{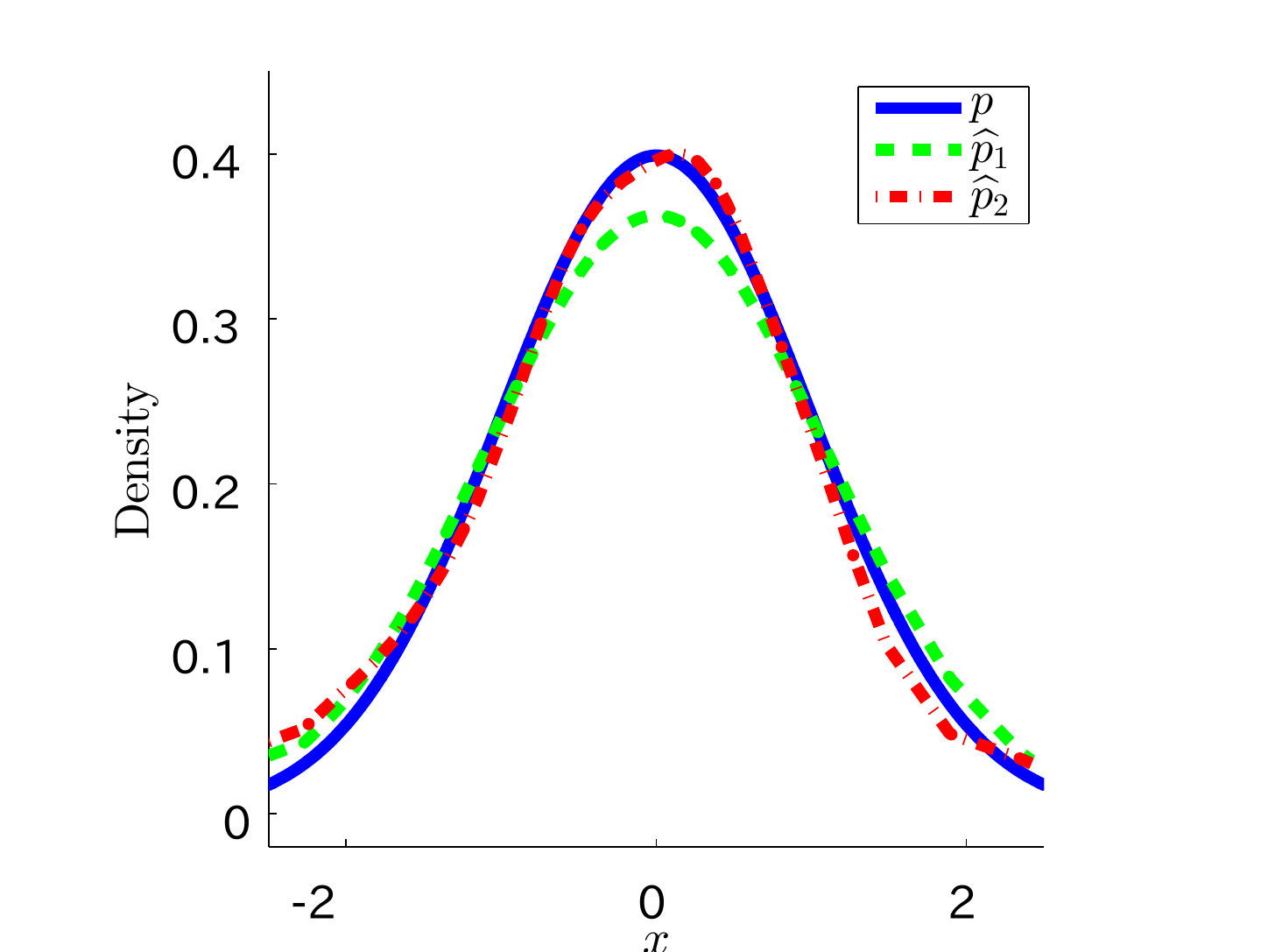}
    \caption{Density estimation}
  \end{subfigure}
  \begin{subfigure}{0.45\textwidth}
    \centering
    \includegraphics[width=\hsize,clip]{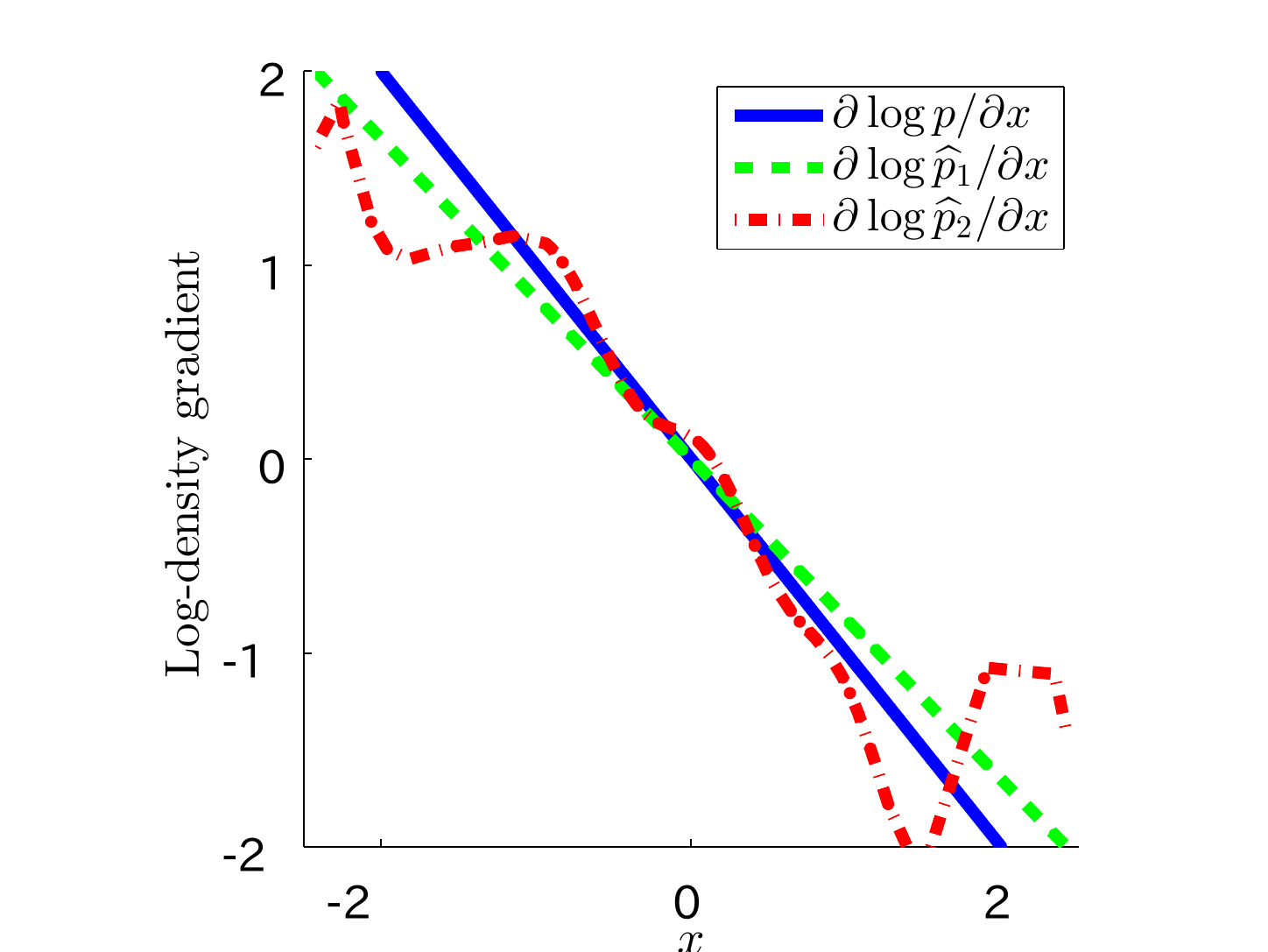}
    \caption{Log-density gradient estimation}
  \end{subfigure}
 \caption{\label{fig:illustration} A comparison of two log-density
 gradient estimates based on density estimation. In (a), $\widehat{p}_2$
 is a better estimate to the true density $p$ than $\widehat{p}_1$,
 while in (b), $\nabla\log\widehat{p}_1$ is a better estimate to the
 true log-density gradient $\nabla\log p$ than
 $\nabla\log\widehat{p}_2$.}
\end{figure*}
Suppose that we are given a set of samples, $\{\bm x_i\}_{i=1}^n$, which
are independent and identically distributed from a probability
distribution with unknown density $p(\bm x)$ on $\Re^d$.  The problem is to
estimate the gradient of the logarithm of the density $p(\bm x)$ from
$\{\bm x_i\}_{i=1}^n$:
\begin{align*}
\nabla\log p(\bm x) &=\left(\partial_1\log p(\bm x), \dotsc,
\partial_d\log p(\bm x)\right)^\top =\left(\frac{\partial_1 p(\bm
x)}{p(\bm x)}, \dotsc, \frac{\partial_d p(\bm x)}{p(\bm x)}\right)^\top,
\end{align*}
where $\partial_j$ denotes the partial derivative operator
$\partial/\partial x^{(j)}$ for $\bm x = (x^{(1)},\dots,x^{(d)})^\top$.

A naive method for estimating the log-density gradient is to first
estimate the probability density, which is performed by, e.g., kernel density estimation (KDE) as
\begin{align*}
\widehat{p}(\bm x) = \frac{1}{n}\sum_{i=1}^n
\frac{1}{\lr(){2\pi\sigma^2}^{\frac{d}{2}}} \exp\lr(){-\frac{\norm{\bm x
- \bm x_i}^2}{2\sigma^2}},
\end{align*}
where $\sigma > 0$ denotes the Gaussian bandwidth, then to take the
gradient of the logarithm of $\widehat{p}(\bm x)$ as
\begin{align*}
\partial_j\log \widehat{p}(\bm x) =\frac{\partial_j \widehat{p}(\bm
x)}{\widehat{p}(\bm x)}.
\end{align*}

However, this two-step method does not work well because an accurate
density estimate does not necessarily provide an accurate log-density
gradient estimate. For example, Figure~\ref{fig:illustration}
illustrates that a worse (or better) density estimate can produce a
better (or worse) gradient estimate.

To overcome this problem, LSLDG, a single-step method
which directly estimates the gradient without going through density estimation,
was proposed~\citep{Beran76,Cox85,Sasaki14},
and has been demonstrated to experimentally work well.
Next, we review LSLDG\@.
% and discuss its potential drawback.
%
\subsection{Direct estimation of log-density gradients}
\label{sec:LSLDG}
The basic idea of LSLDG is to directly fit a model $g_j(\bm x)$ to the
true log-density gradient $\partial_j\log p(\bm x)$ under the squared
loss:
\begin{align*}
 R_j(g_j) &:= \int\lr(){g_j(\bm x)-\partial_j\log p(\bm{x})}^2 p(\bm
 x)\dx\\ &\phantom{:}= \int\lr(){g_j(\bm x)-\frac{\partial_j p(\bm
 x)}{p(\bm x)}}^2 p(\bm x)\dx\\ &\phantom{:}= \int g_j(\bm x)^2 p(\bm
 x)\dx - 2\int g_j(\bm x) \partial_j p(\bm x)\dx + C_j\\ &\phantom{:}=
 \int g_j(\bm x)^2 p(\bm x)\dx -2\int \big[g_j(\bm x)p(\bm
 x)\big]_{x^{(j)}=-\infty}^{x^{(j)}=\infty} \dx^{(\setminus j)} + 2\int
 \partial_j g_j(\bm x) p(\bm x)\dx + C_j\\ &\phantom{:}= \int g_j(\bm
 x)^2 p(\bm x)\dx + 2\int \partial_j g_j(\bm x) p(\bm x)\dx + C_j,
\end{align*}
where $C_j := \int\lr(){\partial_j p(\bm x)}^2 p(\bm x)\dx$ is a
constant that does not depend on $g_j$, $\int(\cdot)\dx^{({\setminus j})}$ denotes
integration except for $x^{(j)}$, and the last deformation comes from
\emph{integration by parts} under the mild condition that $g_j(\bm
x)p(\bm x) \to 0$ as $|x^{(j)}| \rightarrow \infty$.

Then, the \emph{LSLDG score} $J_j(g_j)$ is given as an empirical
approximation to the risk $R_j(g_j)$ subtracted by $C_j$:
\begin{align}
 J_j(g_j) :=\frac{1}{n}\sum_{i=1}^n g_j(\bm x_i)^2 +
 \frac{2}{n}\sum_{i=1}^n \partial_j g_j(\bm
 x_i).\label{eq:LSLDGcriterion}
\end{align}

As $g_j(\bm{x})$, a linear-in-parameter model is used:
\begin{align}
 g_j(\bm x) = \bm\theta_j^\top\bm\psi_j(\bm x) = \sum_{k=1}^b
 \theta_j^{(k)}\psi_j^{(k)}(\bm x),\label{eq:linearmodel}
\end{align}
where $\theta_j^{(k)}$ is a parameter, $\psi_j^{(k)}(\bm x)$ is a
differentiable basis function, and $b$ is the number of the basis
functions.
By substituting \eqref{eq:linearmodel} into
\eqref{eq:LSLDGcriterion} and adding an $\ell_2$-regularizer, we can
analytically obtain the optimal solution $\widehat{\bm\theta}_j$ as
\begin{align*}
\widehat{\bm\theta}_j &=\arg\min_{\bm\theta_j}\lr[]{\bm\theta_j^\top\bm
G_j\bm\theta_j+2\bm h_j^\top\bm\theta_j+\lambda_j\norm{\bm\theta_j}^2}\\
&= -(\bm G_j + \lambda_j\bm I_b)^{-1}\bm h_j,
\end{align*}
where $\lambda_j\ge 0$ is the regularization parameter, $\bm I_b$ is the
$b\times b$ identity matrix, and
\begin{align*}
\bm G_j := \frac{1}{n}\sum_{i=1}^n\bm\psi_j(\bm x_i)\bm\psi_j(\bm
x_i)^\top, \ \bm h_j := \frac{1}{n}\sum_{i=1}^n \partial_j \bm\psi_j(\bm
x_i).
\end{align*}
Finally, an estimator of the log-density gradient is obtained by
\begin{align*}
\hat g_j(\bm x):=\hat{\bm\theta}_j^\top\bm\psi_j(\bm x).
\end{align*}

It was experimentally shown that LSLDG produces much more accurate
estimates of log-density gradients than the KDE-based gradient
estimator and that the clustering method based on LSLDG performs
well~\citep{Sasaki14}.

%3-----------------------------------------
\section{Regularized multi-task learning}
\label{sec:RMTL}
In this section, we review a multi-task learning framework called
\emph{regularized multi-task
learning}~\citep{RMTL,evgeniou2005learning}, which is powerful and
widely applicable to many machine learning methods.

Consider that we have $T$ tasks of supervised learning as follows.  The
task~$t$ is to learn an unknown function $f_t^*(\bm x)$ from samples of
input-output pairs $\{(\bm x_i^{(t)}, y_i^{(t)})\}_{i=1}^{n_t}$, where
$y_i^{(t)}$ is the output $f_t^*(\bm x)$ with noise at the input $\bm x
= \bm x_i^{(t)}$.  When $f_t^*(\bm x)$ is modeled by a parameterized function
$f_t(\bm x; \bm\theta_t)$,  learning is performed by finding the
parameter $\bm\theta_t$ which minimizes the empirical risk associated with
some loss function $l(y,y')$:
\begin{align*}
  \widehat{\bm\theta}_t
  &= \arg\min_{\bm\theta_t} \frac{1}{n_t} \sum_{i=1}^{n_t} l(y_i^{(t)},f_t(\bm x_t^{(t)};\bm\theta_t)) = \arg\min_{\bm\theta_t} J_t(\bm\theta_t),
\end{align*}
where $J_t(\bm\theta_t) = \sum_{i=1}^{n_t} l(y_i^{(t)},f_t(\bm
x_t^{(t)};\bm\theta_t))$.

In regularized multi-task learning, the objective function has regularization terms which impose every pair of parameters to be close to each other while $J_t(\bm\theta_t)$ are jointly minimized:
\begin{align*}
\sum_{t=1}^T J_t(\bm\theta_t)
+ \frac{1}{2}\gamma\sum_{t=1,t'=1}^T
\gamma_{t,t'}\norm{\bm\theta_t-\bm\theta_{t'}}^2,
\end{align*}
where $\gamma\ge 0$ is the regularization parameter and $\gamma_{t,t'}\ge 0$ are the similarity parameters between the tasks $t$ and $t'$.

%\citet{evgeniou2005learning} showed that regularized multi-task learning can be translated into a traditional kernel method with some specific kernel function,
It was experimentally demonstrated that the \emph{multi-task support vector regression}~\citep{RMTL,evgeniou2005learning}, performs better than the single-task counterpart~\citep{vapnik1997svregress} especially when the tasks are highly related each other.

%4-----------------------------------------
\section{Proposed method}
\label{sec:proposed}
In this section, we present our proposed method and algorithms.
\subsection{Basic idea}
Our goal in this paper is to improve the performance of LSLDG
in \emph{multi-dimensional} cases.
For multi-dimensional input $\bm x$,
the log-density gradient $\nabla\log p(\bm x)$ has multiple output dimensions,
meaning that its estimation actually consists of multiple learning tasks.
Our basic idea is to apply regularized multi-task learning to solve these tasks
simultaneously instead of learning them independently.

This idea is supported by the fact that the target functions of these tasks,
$\partial_1 \log p(\bm x), \dots, \partial_d \log p(\bm x)$,
are all derived from the same log-density $\log p(\bm x)$,
and thus they must be strongly related to each other.
Under such strong relatedness, jointly learning them with sharing information
with each other would improve estimation accuracy as has been observed
in other existing multi-task learning work.

  \subsection{Regularized multi-task learning for least-squares
  log-density gradients (MT-LSLDG)}
  \label{sec:MT-LSLDG}
  Here, we propose a method called \emph{regularized multi-task learning for least-squares log-density gradients} (MT-LSLDG).
  
  Our method MT-LSLDG is given by applying regularized multi-task learning to LSLDG\@.
  Specifically, we consider the problem of minimizing the following objective function:
  \begin{align}
   J(\bm\theta_1,\dots,\bm\theta_d) &= \sum_{j=1}^d
   J_j(g_j(\cdot;\bm\theta_j)) + \sum_{j=1}^d
   \lambda_j\norm{\bm\theta_j}^2 + \frac{1}{2}\gamma\sum_{j,j'=1}^d
   \gamma_{j,j'}\norm{\bm\theta_j-\bm\theta_{j'}}^2\nonumber\\ &=
   \sum_{j=1}^d \left(\bm\theta_j^\top \bm{G}_j\bm\theta_j
   +2\bm\theta_j^\top\bm h_j +\lambda_j\norm{\bm\theta_j}^2\right)
   +\frac{1}{2}\gamma\sum_{j,j'=1}^d\gamma_{j,j'}\norm{\bm\theta_j-\bm\theta_{j'}}^2,
   \label{eq:objective}
  \end{align}
  where the last term is the multi-task regularizer which imposes the parameters close to each other.

  Denoting the minimizers of (\ref{eq:objective}) by
  $\bm{\widehat{\theta}}_1,\dots,\bm{\widehat{\theta}}_d$, the estimator
  $\widehat{\bm{g}}(\bm x) =
  (\widehat{g}_1(\bm{x}),\dots,\widehat{g}_d(\bm{x}))^\top$ is given by
  \begin{align}
   \widehat{g}_j(\bm{x})=
   g_j(\bm{x};\widehat{\bm\theta}_j)=\widehat{\bm\theta}_j^\top
   \bm\psi_j(\bm x). \label{eqn:MTLSLDG}
  \end{align}
  We call this method \emph{regularized multi-task learning for
  least-squares log-density gradients} (MT-LSLDG).
  
  \subsection{The design of the basis functions}
  \label{sec:basisfunc}
  The design of the basis functions $\bm\psi_j(\bm x)$ in MT-LSLDG is
  crucial to enjoy the advantage of regularized multi-task learning.  A simple
  design would be to use a common function $\bm\phi(\bm
  x)=(\phi^{(1)}(\bm x),\dots,\phi^{(b)}(\bm x))$ to all $\bm\psi_j(\bm
  x)$, that is, $\bm\psi_1(\bm x) = \dots =
  \bm\psi_d(\bm x) = \bm\phi(\bm x)$.  From (\ref{eq:objective}) and (\ref{eqn:MTLSLDG}),
  in this design, the multi-task regularizer promotes $\widehat{g}_j(\bm
  x;\widehat{\bm\theta}_j)$ to be more close to each other so that
  \begin{align*}
   \widehat{g}_1(\bm x;\widehat{\bm\theta}_1) \approx \dots \approx
   \widehat{g}_d(\bm x;\widehat{\bm\theta}_d).
  \end{align*}
  However, it is inappropriate that all $\widehat{g}_j(\bm
  x;\widehat{\bm\theta}_j)$ are similar because the different true
  partial derivatives, say $\partial_j\log p(\bm x)$ and
  $\partial_{j^{\prime}}\log p(\bm x)$ for $j\neq j^{\prime}$, show
  different profiles in general.
  
  To avoid this problem, we propose to use the partial derivatives of
  $\bm\phi(\bm x)$ as basis functions:
  \begin{align*}
   \bm \psi_j^{(k)}(\bm x) = \partial_j \bm\phi(\bm x).
  \end{align*}
  For this basis function design, it holds that
  \begin{align*}
   \widehat{\bm g}_j(\bm x) =\widehat{\bm\theta}_j^\top\bm \psi_j(\bm x)
   =\widehat{\bm\theta}_j^\top\partial_j\bm\phi(\bm x)
   =\partial_j\widehat{\bm\theta}_j^\top\bm\phi(\bm x) \approx
   \partial_j\log p(\bm x).
  \end{align*}
  Thus, $\widehat{\bm\theta}_j^\top\bm\phi(\bm x)$ are approximations of the
  true log-density $\log p(\bm x)$.
  Since the multi-task regularizer encourages the log-density estimates
  $\widehat{\bm\theta}_j^\top\bm\phi(\bm x)$ to be more similar,
  this basis design would be reasonable.
  
  As a specific choice of $\phi^{(k)}(\bm x)$, we use a Gaussian kernel:
  \begin{align*}
   \psi_j^{(k)}(\bm x) &= \partial_j \exp\lr(){-\frac{\norm{\bm x - \bm
   c_k}^2}{2\sigma_j^2}}\\ &= \frac{c_k^{(j)} -
   x^{(j)}}{\sigma_j^2}\exp\lr(){-\frac{\norm{\bm x - \bm
   c_k}^2}{2\sigma_j^2}},
  \end{align*}
  where $\bm c_k$ are the centers of the kernels, and $\sigma_j > 0$ are
  the Gaussian band width parameters.  

  \subsection{Hyper-parameter selection by cross-validation}
  \label{ssec:CV}
  As in LSLDG, the hyper parameters, which are the
  $\ell_2$-regularization parameters $\lambda_j$, the Gaussian width
  $\sigma_j$, and the multi-task parameters $\gamma, \gamma_{j,j'}$, can
  be cross-validated in MT-LSLDG. The procedure of the $K$-fold
  cross-validation is as follows: First, we randomly partition the set
  of training samples $S_\text{tr}$ into $K$ folds
  $F_1,\dots,F_K$. Next, for each $k=1,\dots,K$, we estimate the
  log-density gradient using the samples in $S_\text{tr}\setminus F_k$
  , which is denoted by $\hat g_j^{(k)}$, and then calculate the LSLDG
  scores for the samples in $F_k$ as $J_\text{CV}^{(k)}$:
 \begin{align*}
  J_\text{CV}^{(k)}
  =\frac{1}{\abs{F_k}}\sum_{\bm x\in F_k} \widehat g_j^{(k)}(\bm x)^2
  + \frac{2}{\abs{F_k}}\sum_{\bm x\in F_k} \frac{\partial \widehat g_j^{(k)}(\bm x)}{\partial x^{(j)}}.
 \end{align*}
 We average these LSLDG scores to obtain the $K$-fold cross-validated
 LSLDG score:
 \begin{align*}
  J_\text{CV} = \frac{1}{K}\sum_{k=1}^K J_\text{CV}^{(k)}.
 \end{align*}
 Finally, we choose the hyper-parameters that minimize $J_\text{CV}$.
 Throughout this paper, we set $K = 5$.
  \subsection{Optimization algorithms in MT-LSLDG}
  Here, we develop two algorithms for minimizing \eqref{eq:objective}.
  One algorithm is to directly evaluate the analytic solution and the other is an
  iterative method based on block coordinate
  descent~\citep{warga1963minimizing}.
   \subsubsection{Analytic solution}
   For simplicity, we assume the similarity parameters are symmetric:
   $\gamma_{j,j'} = \gamma_{j',j}$.  Then, the objective function
   $J(\bm\theta_1,\dots,\bm\theta_d)$ can be expressed as a quadratic
   function in terms of $\bm\theta =
   (\bm\theta_1^\top,\dots,\bm\theta_d^\top)^\top$ as
   \begin{align*}
    J(\bm\theta) &= \bm\theta^\top(\bm G
    +\bm C\otimes\bm I_b)\bm\theta+2\bm\theta^\top\bm h,
   \end{align*}
   where
   \begin{align*}
    \bm G &= \diag(\bm G_1,\dots,\bm G_d),\ \bm h = (\bm
    h_1^\top,\dots,\bm h_d^\top)^\top,\\ \bm C &:=
    \diag(\lambda_1,\dots,\lambda_d) +
    \gamma\diag\lr(){\sum_{j=1}^d\gamma_{1,j},\dots,\sum_{j=1}^d\gamma_{d,j}}
    - \gamma\bm\Gamma,
   \end{align*}
   $[\bm\Gamma]_{j,j'}=\gamma_{j,j'}$, $\diag(\cdot,\dots,\cdot)$ is
   a block-diagonal matrix whose diagonal blocks are its arguments, and
   $\otimes$ denotes the Kronecker product.  The minimizer
   $\widehat{\bm\theta}$ of $J(\bm\theta)$ is analytically computed by
   \begin{align}
    \widehat{\bm\theta}&=\arg\min_{\bm\theta}J(\bm\theta) =-(\bm G+\bm
    C\otimes\bm I_b)^{-1}\bm h.\label{eq:analyticMTLLSLDG}
   \end{align}

   \subsubsection{Block coordinate descent (BCD) method}
	\label{sssec:BCD}
   Direct computation of the analytic solution
   \eqref{eq:analyticMTLLSLDG} involves inversion of a $db\times db$
   matrix.  This may be not only expensive in terms of computation time
   but also infeasible in terms of memory space when the dimensionality
   $d$ is very large.

   Alternatively, we propose an algorithm based on
   block coordinate descent (BCD)~\citep{warga1963minimizing}.
   It is an iterative algorithm
   which only needs manipulation of a relatively small $b\times b$
   matrix at each iteration. This alleviates the memory size requirement
   and hopefully reduces computation time if the number of iterations is
   not large.

   A pseudo code of the algorithm is shown in
   Algorithm~\ref{alg:BCD-MTL}.  At each update \eqref{eq:BCDUpdate} in
   the algorithm, only one vector $\widehat{\bm\theta}_j$ is optimized
   in a closed-form while fixing the other parameters
   $\widehat{\bm\theta}_{j'}$ ($j'\neq j$).  The update
   \eqref{eq:BCDUpdate} only requires computing the inverse of a
   $b\times b$ matrix, which seems to be computationally advantageous over evaluating
   the analytic solution in terms of the computation cost and memory
   size requirement.

{\newcommand{\tbtheta}{\widetilde{\bm\theta}}
	Another important technique to reduce the overall computation time is
to use \emph{warm start} initialization:
	when the optimal value of $\gamma$ is searched for by cross-validation,
	we	may use the solutions $\tbtheta_1,\dots,\tbtheta_d$ obtained
	with $\gamma$ as initial values for another $\gamma$.

\begin{algorithm}[t]
\caption{Block coordinate descent (BCD) algorithm.}
\label{alg:BCD-MTL}
\begin{algorithmic}
  \State Initialize $\tbtheta_1,\dots,\tbtheta_d$.
  \Repeat
    \For{$j=1,\dots,d$}
      \vspace*{-1.4cm}
      \State \begin{align}
      \widetilde{\bm\theta}_j &\gets \arg\min_{\bm\theta_j}
      J(\widetilde{\bm\theta}_1,\dotsc,\widetilde{\bm\theta}_{j-1},\bm\theta_j,\widetilde{\bm\theta}_{j+1},\dotsc,\widetilde{\bm\theta}_d)\nonumber\\
      &= \lr(){\bm G_j+\lambda_j\bm I_b+2\gamma\sum_{j\neq j'}\gamma_{j,j'}\bm I_b}^{-1}
      \lr(){-\bm h_j+2\gamma\sum_{j'\neq j}\gamma_{j,j'}\widetilde{\bm\theta}_{j'}}. \label{eq:BCDUpdate}
      \end{align}
    \EndFor
  \Until{$\tbtheta_1,\dots,\tbtheta_d$ converge.}
\end{algorithmic}
\end{algorithm}
}

%6-----------------------------------------
\section{Experiments on log-density gradient estimation}
\label{sec:gradExperiments}
In this section, we illustrate the behavior of the proposed method and
experimentally investigate its performance.

\subsection{Experimental setting}
In each experiment, training samples $\{\bm x_i\}_{i=1}^n$ and test
samples $\{\bm x'_{i'}\}_{i'=1}^{n'}$ are drawn independently
from an unknown density $p(\bm x)$.
We estimate $\nabla\log p(\bm x)$ from the training samples,
and then evaluate the estimation performance by the \emph{test score}
\begin{align*}
  J_\text{te}(\widehat{\bm g}) = \sum_{j=1}^d \left[\frac{1}{n'}\sum_{i'=1}^{n'} \widehat{g}_j(\bm x'_{i'})^2
  +\frac{2}{n'}\sum_{i'=1}^{n'}\partial _j \widehat{g}_j(\bm x'_{i'})\right],
\end{align*}
where $\widehat{\bm g}(\bm x) = (\widehat{g}_1(\bm
x),\dots,\widehat{g}_d(\bm x))^\top$ is an estimated log-density
gradient.  This score is an empirical approximation of the expected
squared loss of $\widehat{\bm g}(\bm x)$ over the test samples
without the constant $C_j$ (see Section~\ref{sec:LSLDG}), and a smaller
score means a better estimate.

We compare the following three methods:
\begin{itemize}
 \item The multi-task LSLDG (MT-LSLDG): our method proposed in
       Section~\ref{sec:proposed}.

 \item The single-task LSLDG (S-LSLDG): the existing
       method~\citep{Beran76,Cox85} reviewed in
       Section~\ref{sec:LSLDG}. This method agrees with MT-LSLDG at
       $\gamma = 0$.
       
 \item The common-parameter LSLDG (C-LSLDG): LSLDG with common
       parameters $\bm\theta' = \bm\theta_1 = \dots = \bm\theta_d$
       learned simultaneously. The solution is given as
       \begin{align*}
	\widehat{\bm\theta'} & = \arg\min_{\bm\theta'}
	\lr[]{\bm\theta'^\top\sum_{j=1}^d\bm G_j\bm\theta' +
	2\sum_{j=1}^d\bm h_j^\top\bm\theta' +
	\lambda\norm{\bm\theta'}^2}\\ & = -\lr(){\sum_{j=1}^d\bm G_j +
	\lambda\bm I_b}^{-1}\sum_{j=1}^d\bm h_j,
       \end{align*}
       where $\lambda\ge 0$ is the $\ell_2$-regularization parameter.  This
       method agrees with MT-LSLDG at the limit $\gamma \to \infty$.
\end{itemize}

In all the methods, we set the number of basis functions as
$b=\min\{50,n\}$, and randomly choose the kernel centers $\bm c_1,
\dots, \bm c_b$ from training samples $\{\bm x_i\}_{i=1}^n$. For
hyper-parameters, we use the common $\ell_2$-regularization parameter
$\lambda$ and bandwidth parameter $\sigma$ among all the dimensions,
$\lambda_1 = \dots = \lambda_d = \lambda$ and $\sigma_1 = \dots = \sigma_d =
\sigma$.  We also set all the similarity parameters as $\gamma_{j,j'} =
1$, which assumes that all dimensions are equally related to each
other.

  \subsection{Artificial data}
  We conduct numerical experiments on artificial data to investigate the
  basic behavior of MT-LSLDG\@.
  As data density $p(\bm x)$, we consider the
  following two cases:
  \begin{itemize}
   \item Single Gaussian: The $d$-dimensional Gaussian density whose
	 mean is $\bm 0$ and whose covariance matrix is the diagonal
	 matrix with the first half of the diagonal elements are $1$ and
	 the others are $5$.
   \item Double Gaussian: A mixture of two $d$-dimensional Gaussian
	 densities with mean zero and $(5,0,\dots,0)^\top$ and
   identity covariance matrix.
   The mixing coefficients are $1/2$.
  \end{itemize}
  The dimensionality $d$ and sample size $n$ are specified later.
  
   %\subsubsection{Does MT-LSLDG improve the original LSLDG?}
   %
   \begin{figure*}[t]
    \centering
    \begin{subfigure}[b]{0.45\columnwidth}
     \includegraphics[width=\hsize,clip]{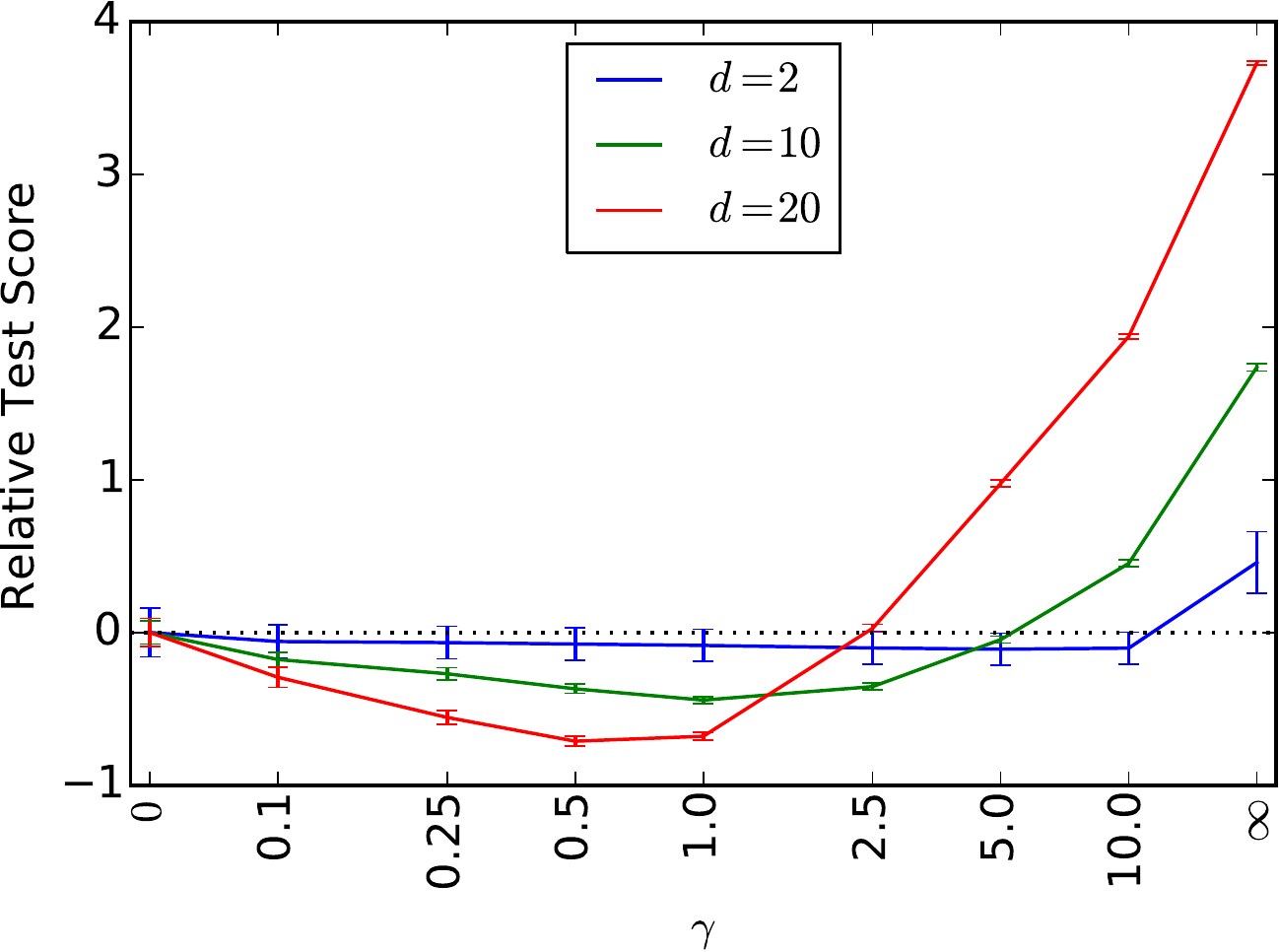}
     \caption{Single Gaussian ($n=30$)}\label{fig:singleGauss_dims}
    \end{subfigure}
    \begin{subfigure}[b]{0.45\columnwidth}
     \includegraphics[width=\hsize,clip]{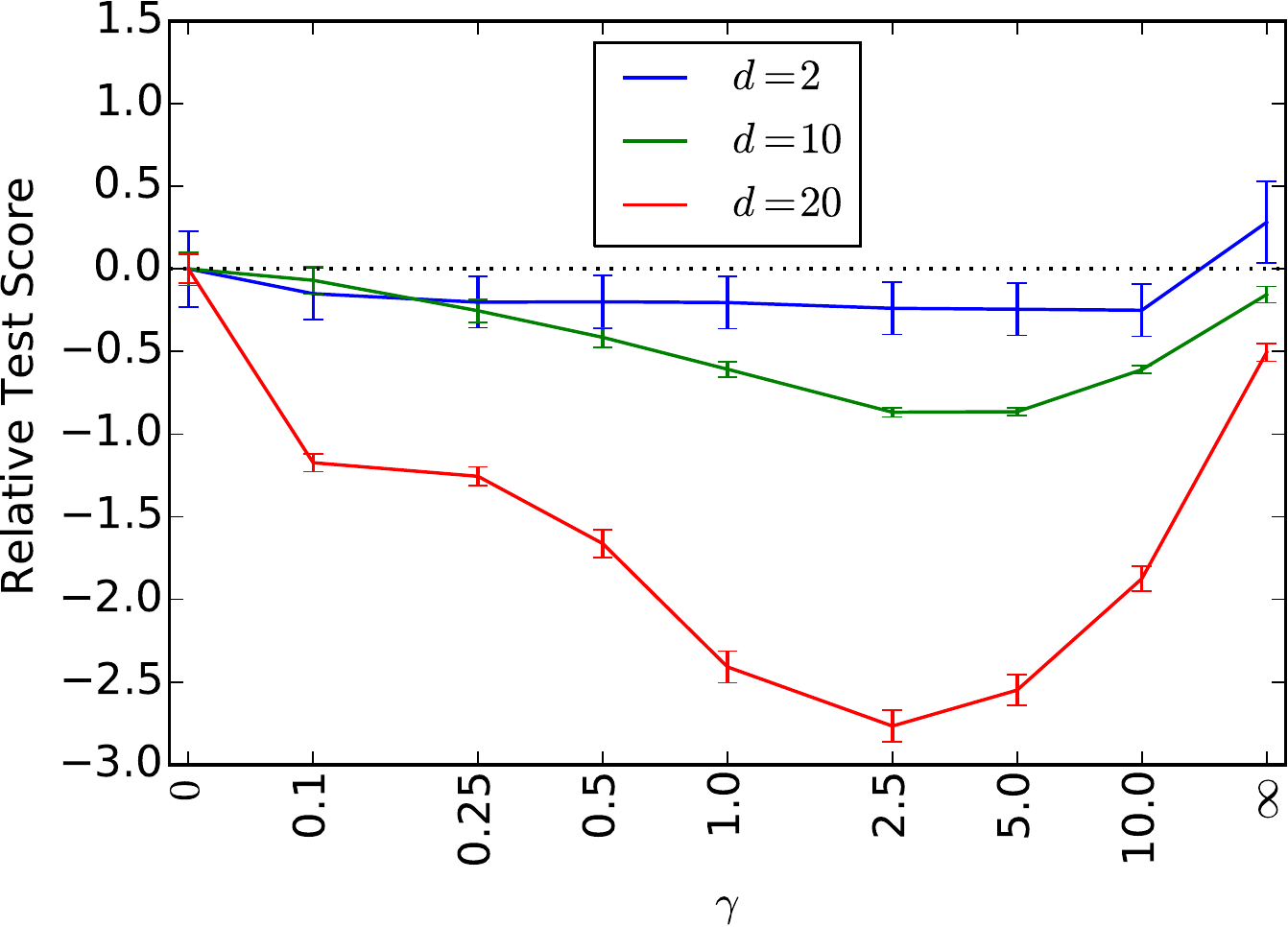}
     \caption{Double Gaussian ($n=30$)}\label{fig:doubleGauss_dims}
    \end{subfigure}
    \begin{subfigure}[b]{0.45\columnwidth}
    \centering \includegraphics[width=\hsize,clip]{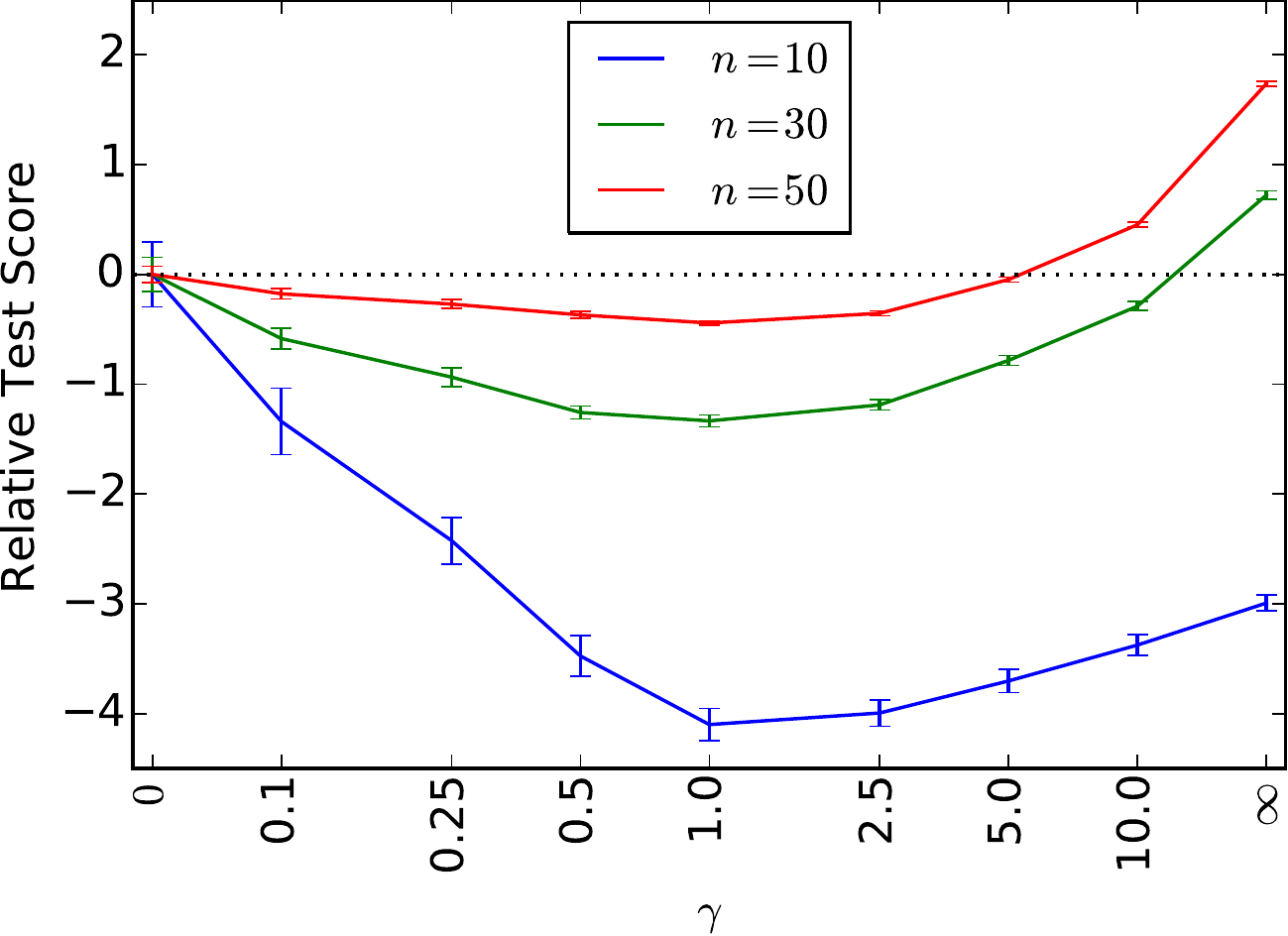}
    \caption{Single Gaussian ($d=10$)}\label{fig:singleGauss_ns}
    \end{subfigure}
    \begin{subfigure}[b]{0.45\columnwidth}
     \includegraphics[width=\hsize,clip]{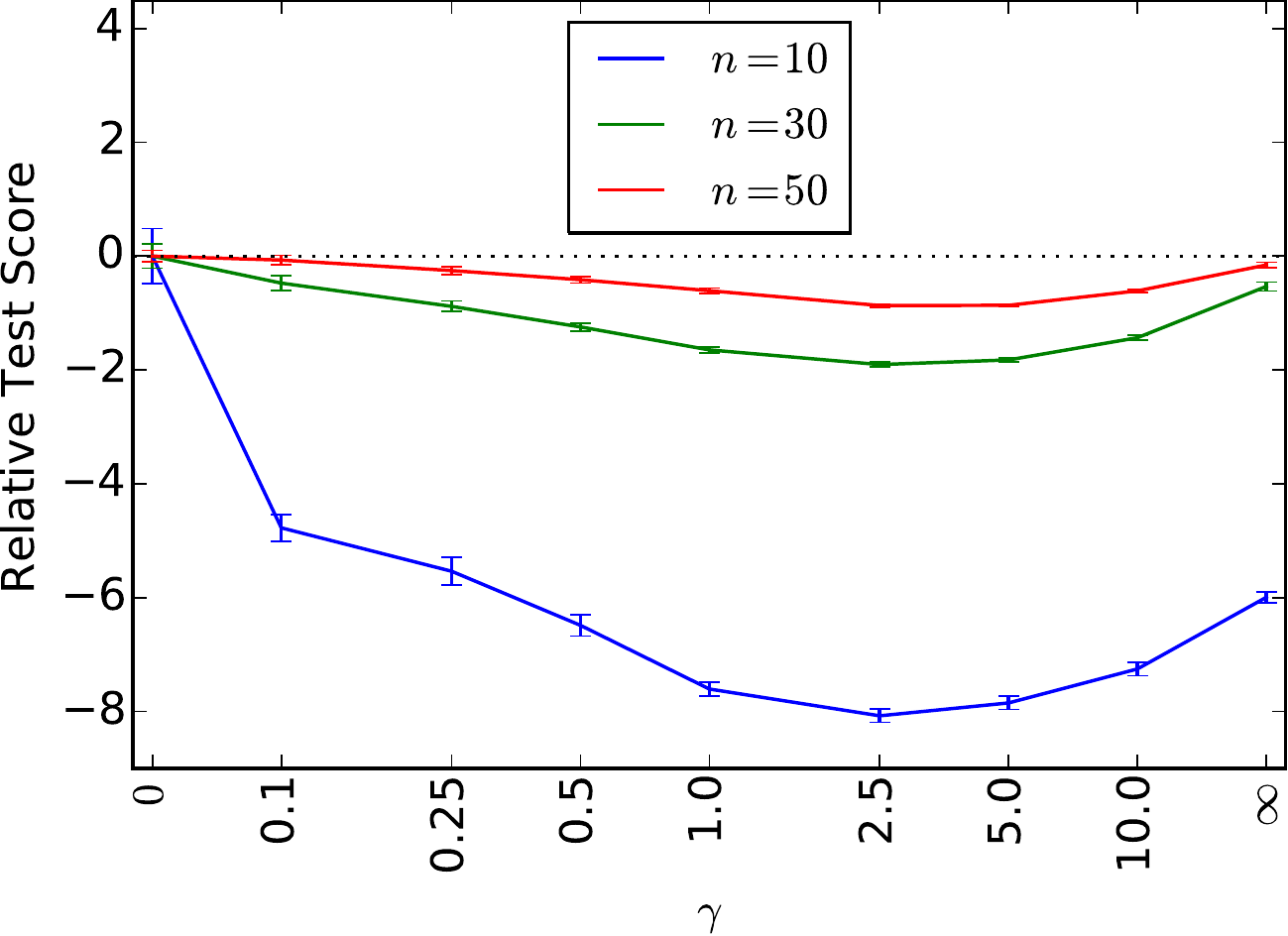} \caption{Double
    Gaussian ($d=10$)}\label{fig:doubleGauss_ns}
    \end{subfigure}
    \caption{Average (and standard errors) of \emph{relative test
    scores} over 100 runs.  The relative test scores refer to test
    scores from which the test score of S-LSLDG is subtracted.
    The black dotted lines indicate the relative score zero.}
    \label{fig:score_vs_gamma}
   \end{figure*}
   First, we investigate whether MT-LSLDG improves the estimation
   accuracy of LSLDG at appropriate $\gamma$.
   We prepare datasets with different dimensionalities $d=2,10,20$ and
   sample sizes $n=10,30,50$.  MT-LSLDG is applied to the datasets at
   each $\gamma\in\{0,0.1,0.25,0.5,1,2.5,5,10,\infty\}$. The Gaussian
   bandwidth $\sigma$ and the $\ell_2$-regularization parameter
   $\lambda$ are chosen by $5$-fold cross-validation as described in
   Section~\ref{sec:proposed} from the candidate lists
   $\{10^{-1},10^{-0.25},10^{0.5},10^{1.25},10^2\}$ and
   $\{10^{-2},10^{-1.25},10^{0.5},10^{0.25},10^1\}$, respectively. The
   solution of MT-LSLDG is computed analytically as in
   \eqref{eq:analyticMTLLSLDG}.
   
   The results are plotted in Figure~\ref{fig:score_vs_gamma}. In the
   figure, the relative test score is defined as the test score
   from which the test score of S-LSLDG is subtracted,
   and thus negative relative scores indicate that
   MT-LSLDG improved the performance of S-LSLDG\@.  When $d=2$, MT-LSLDG
   does not improve the performance for any $\gamma$ values
   (Figure~\ref{fig:singleGauss_dims}
   and~\ref{fig:doubleGauss_dims}). However, for higher-dimensional
   data, the performance is improved at appropriate
   $\gamma$ values (e.g., $\gamma = 0.5$ for $d=20$ in
   Figure~\ref{fig:singleGauss_dims} and $\gamma = 2.5$ for $d=20$ in
   Figure~\ref{fig:doubleGauss_dims}).  Similar improvement is observed
   also for smaller sample size (e.g., $n=10$ and $n
   = 30$) in Figure~\ref{fig:singleGauss_ns} and
   Figure~\ref{fig:doubleGauss_ns}.
      
   These results confirm that MT-LSLDG improves the performance of
   S-LSLDG at an appropriate $\gamma$ value when data is
   relatively high-dimensional and the sample size is small. Since such $\gamma$ is
   usually unknown in advance, we need to find a reasonable value in
   practice.
   %
   %\subsubsection{The multi-task regularization parameter selection by
   %cross-validation}

   Next, we investigate whether an appropriate $\gamma$ value can be
   chosen by cross-validation. In this experiment, the cross-validation
   method in Section~\ref{ssec:CV} is performed to choose $\gamma$ as
   well. The candidates of $\gamma$ is
   $\{0,0.1,0.25,0.5,1,2.5,5,10,\infty\}$. The other experimental
   settings such as the data generation and all the LSLDGs are the same
   as in the last experiment.
   
   Table~\ref{table:testWithCV} shows that MT-LSLDG improves the performance
   especially when the dimensionality of data is relatively high and the sample
   size is small. These results indicate that the proposed
   cross-validation method allows us to choose a reasonable $\gamma$
   value.

   \begin{table}[t]
   \centering
   \caption{Averages (and standard errors) of test scores on the artificial data with cross-validation over 100 runs.
   In each row, the best and comparable to the best scores in terms of paired t-test
   with significance level $5 \%$ are emphasized in bold face.}
	\vspace{3mm}
   \label{table:testWithCV}
   \begin{tabular}{lrrrr}
   \hline
   Density & $n$ & MT-LSLDG & S-LSLDG & C-LSLDG\\
   \hline
   \hline
   Single Gaussian
   & $10$ & $\bm{-2.87}$ ($0.22$) & $0.37$ ($0.31$) & $\bm{-2.58}$ ($0.07$)\\
   & $30$ & $\bm{-5.34}$ ($0.038$) & $-4.97$ ($0.08$) & $-3.29$ ($0.03$)\\
   & $50$ & $\bm{-5.63}$ ($0.02$) & $-5.55$ ($0.02$) & $-4.13$ ($0.04$)\\
   \hline
   Double Gaussian
   & $10$ & $\bm{-6.83}$ ($0.14$) & $1.01$ ($0.54$) & $-5.02$ ($0.12$)\\
   & $30$ & $\bm{-8.45}$ ($0.03$) & $-7.63$ ($0.10$) & $-7.84$ ($0.04$)\\
   & $50$ & $\bm{-8.67}$ ($0.02$) & $-8.29$ ($0.10$) & $-8.48$ ($0.02$)\\
   \hline
   \end{tabular}

   \vspace{3mm}

   \begin{tabular}{lrrrr}
   \hline
   Density & $d$ & MT-LSLDG & S-LSLDG & C-LSLDG\\
   \hline
   \hline
   Single Gaussian
   & $2$  & $\bm{0.20}$ ($0.19$) & $\bm{-0.11}$ ($0.15$) & $\bm{0.09}$ ($0.15$)\\
   & $10$ & $\bm{-5.34}$ ($0.04$) & $-4.97$ ($0.08$) & $-3.29$ ($0.03$)\\
   & $20$ & $\bm{-10.77}$ ($0.03$) & $-9.98$ ($0.13$) & $-6.39$ ($0.01$)\\
   \hline
   Double Gaussian
   & $2$  & $\bm{0.54}$ ($0.22$) & $\bm{0.50}$ ($0.27$) & $\bm{0.19}$ ($0.22$)\\
   & $10$ & $\bm{-8.45}$ ($0.03$) & $-7.63$ ($0.10$) & $-7.84$ ($0.04$)\\
   & $20$ & $\bm{-16.9}$ ($0.14$) & $-14.90$ ($0.10$) & $-15.26$ ($0.06$)\\
   \hline
   \end{tabular}
   \end{table}

   \subsection{Benchmark data}
   In this section, we demonstrate the usefulness of MT-LSLDG in
   gradient estimation on various benchmark datasets.

   This experiment uses IDA benchmark repository~\citep{Ratsch}. For
   MT-LSLDG, the hyper-parameters $\sigma$, $\lambda$ and $\gamma$ are
   chosen by cross-validation. The candidate lists are
   $\sigma\in\{0.1,0.25,0.5,1,2.5,5,10\}$,
   $\lambda\in\{10^{-5},10^{-4},\dots,10^{-1}\}$ and
   $\gamma\in\{0,10^{-5},10^{-4},\dots,10^{1},10^{2},\infty\}$,
   respectively. For S-LSLDG and C-LSLDG, the candidate lists of $\sigma$ and
   $\lambda$ are the same as MT-LSLDG\@. The solution of MT-LSLDG is
   computed by the BCD algorithm described in Section~\ref{sssec:BCD}.
   
   The results are presented in Table~\ref{table:IDAScore}. MT-LSLDG
   significantly improves the performance of either S-LSLDG or C-LSLDG
   on most of the datasets.
   \begin{landscape}
    \begin{table}[t]
      \centering \caption{Avarages (and standard errors) of the test scores on IDA datasets.
      In each dataset, the best and comparable to the best scores in terms of paired t-test with significance level $5 \%$ are emphasized in bold face.
      The number of trials is 20 for the image and splice dataset, and is 100 for the other datasets.}
     \vspace*{5mm}
     \label{table:IDAScore}
     \begin{tabular}{lrrr}
      \hline
      Dataset ($d$, $n$) & MT-LSLDG & S-LSLDG & C-LSLDG\\
      \hline
      \hline
  thyroid ($5$, $140$) & $\bm{-111}$ ($7$) & $\bm{-103}$ ($8$) & $-40$ ($4$)\\
  diabetes ($8$, $468$) & $\bm{-21.2}$ ($3.9$) & $\bm{-18.0}$ ($4.5$) & $\bm{-15.1}$ ($0.1$)\\
  flare-solar ($9$, $666$) & $\bm{-1.4\times10^7}$ ($0.0\times10^7$) & $-0.2\times10^7$ ($0.0\times10^7$) & $-1.4\times10^7$ ($0.0\times10^7$)\\
  breast-cancer ($9$, $200$) & $\bm{-2.6\times10^3}$ ($0.2\times10^3$) & $-0.2\times10^3$ ($0.0\times10^3$) & $\bm{-2.6\times10^3}$ ($0.2\times10^3$)\\
  image ($18$, $1300$) & $\bm{-2.0\times10^4}$ ($0.0\times10^4$) & $\bm{-2.0\times10^4}$ ($0.0\times10^4$) & $-0.0\times10^4$ ($0.0\times10^4$)\\
  german ($20$, $700$) & $\bm{-3.1\times10^2}$ ($0.1\times10^2$) & $\bm{-3.0\times10^2}$ ($0.1\times10^2$) & $-0.2\times10^2$ ($0.0\times10^2$)\\
  twonorm ($20$, $400$) & $\bm{-22.3}$ ($0.0$) & $\bm{-22.3}$ ($0.0$) & $-22.1$ ($0.0$)\\
  waveform ($21$, $400$) & $\bm{-43}$ ($0$) & $-43$ ($0$) & $-35$ ($0$)\\
  splice ($60$, $1000$) & $\bm{-1.8\times10^3}$ ($0.6\times10^3$) & $-0.0\times10^3$ ($0.0\times10^3$) & $\bm{-2.0\times10^3}$ ($0.6\times10^3$)\\
  \hline
  \end{tabular}
  \end{table}
  \end{landscape}

%7-----------------------------------------
\section{Application to mode-seeking clustering}
\label{sec:clustering}
In this section, we apply MT-LSLDG to mode-seeking clustering
and experimentally demonstrate its usefulness.
\subsection{Mode-seeking clustering}
\begin{figure*}[tbhp]
 \centering
 \includegraphics[width=0.6\hsize,clip]{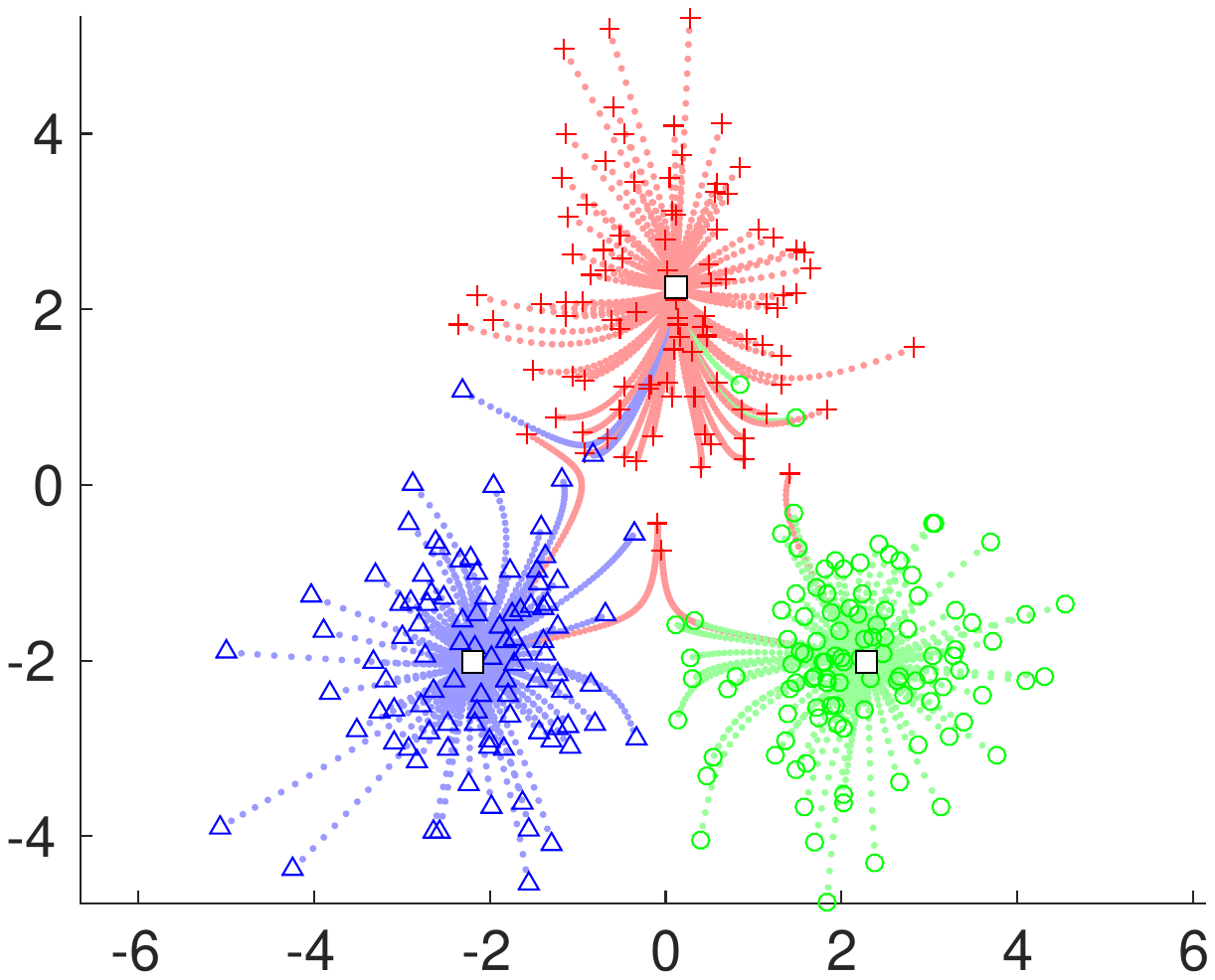}
 \label{fig:threeClusters}
 \caption{\label{fig:clustIllust} Transition of data points during a
 mode-seeking process. Data samples are drawn from a mixture of Gaussians, and
 the data points sampled from the same Gaussian component
 are specified by the same color (red, green, or blue)
 and marker (plus symbol, circle, or triangle).
 White squares indicate the points to which data points converged.}
\end{figure*}
A practical application of log-density gradient estimation is
\emph{mode-seeking clustering}~\citep{fukunaga1975mean, cheng1995mean,
comaniciu2002mean, Sasaki14}. Mode-seeking clustering methods update
each data point toward a nearby mode by gradient ascent,
and assign the same clustering label to the data points which converged to the same
mode (Figure~\ref{fig:clustIllust}). Their notable advantage is that we
need not specify the number of clusters in advance. Mode-seeking
clustering has been successfully applied to a variety of real world problems
such as object tracking~\citep{comaniciu2000real},
image segmentation~\citep{comaniciu2002mean,Sasaki14},
and line edge detection in images~\citep{bandera2006mean}.

In mode-seeking, the essential ingredient is the gradient of the data
density. To estimate the gradients, \emph{mean shift
clustering}~\citep{fukunaga1975mean, cheng1995mean, comaniciu2002mean},
which is one of the most popular mode-seeking clustering methods,
employs the two-step method of first estimating the data density
by kernel density estimation
and then taking its gradient. However, as we mentioned earlier, this
two-step method does not work well since accurately estimating the
density does not necessarily lead to an accurate estimate of the
gradient.

In order to overcome this problem, \emph{LSLDG clustering}~\citep{Sasaki14}
adopted LSLDG instead of the two-step method.
\citet{Sasaki14} also provided a practically useful fixed-point algorithm
for mode-seeking as in mean shift clustering~\citep{cheng1995mean}:
When the partial derivative of a vector of Gaussian kernels
$\bm\psi_j(\bm x) = \partial_j\bm\phi(\bm x)$ is used as the vector of
basis functions, the model $g_j(\bm
x)=\widehat{\bm\theta}_j^\top\bm\psi_j(\bm x)$ can be transformed as
\begin{align*}
 \widehat{g}_j(\bm x) &= \sum_{k=1}^b
 \widehat{\theta}_j^{(k)}\frac{c_k^{(j)}-x^{(j)}}{\sigma^2}\phi^{(k)}(\bm
 x)\\ &= \frac{1}{\sigma^2}\sum_{k=1}^b
 \widehat{\theta}_j^{(k)}c_k^{(j)}\phi^{(k)}(\bm
 x)-\frac{x^{(j)}}{\sigma^2}\sum_{k=1}^b
 \widehat{\theta}_j^{(k)}\phi^{(k)}(\bm x)\\ &=
 \lr[]{\frac{1}{\sigma^2}\sum_{k=1}^b
 \widehat{\theta}_j^{(k)}\phi^{(k)}(\bm x)} \lr[]{\frac{\sum_{k'=1}^b
 \widehat{\theta}_j^{(k')}c_{k'}^{(j)}\phi^{(k')}(\bm
 x)}{\sum_{k'=1}^b\widehat{\theta}_j^{(k')}\phi^{(k')}(\bm x)}-x^{(j)}},
\end{align*}
where we assume that
$\frac{1}{\sigma^2}\sum_{k=1}^b\widehat{\theta}_j^{(k)}\phi^{(k)}(\bm
x)$ is nonzero. Setting $\widehat g_j(\bm x)$ to zero yields a
fixed-point update formula as
\begin{align*}
 x^{(j)} \gets
\frac{\sum_{k'=1}^b\widehat{\theta}_j^{(k')}c_{k'}^{(j)}\phi^{(k')}(\bm
x)}{\sum_{k'=1}^b\widehat{\theta}_j^{(k')}\phi^{(k')}(\bm x)}.
\end{align*}
It has been experimentally shown that LSLDG clustering performs
significantly better than mean-shift clustering~\citep{Sasaki14}.

Here, we apply MT-LSLDG to LSLDG clustering and investigate if the
performance is improved in mode-seeking clustering as well
for relatively high-dimensional data.

\subsection{Experiments}
Next, we conduct numerical experiments for mode-seeking clustering.

\subsubsection{Experimental setting}
We apply the following four clustering methods to various datasets:
\begin{itemize}
 \item MT-LSLDGC: LSLDG clustering with MT-LSLDG.

 \item S-LSLDGC: LSLDG clustering with S-LSLDG~\citep{Sasaki14}.
       
 \item C-LSLDGC: LSLDG clustering with C-LSLDG~\citep{Sasaki14}.

 \item Mean-shift: mean shift clustering~\citep{comaniciu2002mean}.
\end{itemize}
For MTL-, S-, and C-LSLDG, all the hyper-parameters are cross-validated
as described in Section~\ref{ssec:CV},
and for mean-shift, log-likelihood cross-validation is used.

We evaluate the clustering performance by the \emph{adjusted Rand index}
(ARI)~\citep{hubert1985ari}.
ARI gives one to the perfect clustering assignment
and zero on average to a random clustering assignment.
A larger ARI value means a better clustering result.

\subsubsection{Artificial data}
First, we conduct experiments on artificial data.
The density of the artificial data is a mixture of three $d$-dimensional Gaussian densities
with means $(0,2,0,\dots,0)$, $(-2,-2,0,\dots,0)$, and $(2,-2,0,\dots,0)$,
covariance matrices $\tfrac{1}{\sqrt{2\pi}}\bm I_d$,
and mixing coefficients $0.4$,$0.3$,$0.3$.
The candidate lists of the hyper-parameters are the following:
$\sigma\in\{10^{-1},10^{-7/9},10^{-5/9},\dots,10^{5/9},10^{7/9},10^{1}\}$,
$\lambda\in\{10^{-5},10^{-4},10^{-3},10^{-2},10^{-1}\}$ and,
$\gamma\in\{0,10^{-5},10^{-4},10^{-3},10^{-2},10^{-1},10^0,10^1,10^2,\infty\}$.

The results are shown in Table~\ref{table:clusterOnToy}.
We can see that MT-LSLDGC performs well especially for the largest dimensionality $d=20$.

\begin{table}[t]
\caption{Averages (and standard errors) of ARIs on real data.
         In each row, the best and comparable to the best ARI
         in terms of unpaired t-test with significance level $5\%$
         is emphasized in bold face.
         The number of trials is 100.}
 \label{table:clusterOnToy} \centering
 \vspace{3mm}
\begin{tabular}{rcccc}
\hline
$d$ & MT-LSLDGC & S-LSLDGC & C-LSLDGC & Mean-shift\\
\hline
\hline
$2$ & $\bm{0.992}$ ($0.035$) & $\bm{0.973}$ ($0.125$) & $\bm{0.992}$ ($0.036$) & $\bm{0.984}$ ($0.044$)\\
$10$ & $\bm{0.993}$ ($0.004$) & $\bm{0.994}$ ($0.003$) & $\bm{0.994}$ ($0.004$) & $0.042$ ($0.022$)\\
$15$ & $\bm{0.983}$ ($0.023$) & $\bm{0.982}$ ($0.054$) & $0.877$ ($0.217$) & $0.000$ ($0.000$)\\
$20$ & $\bm{0.827}$ ($0.190$) & $0.586$ ($0.208$) & $0.716$ ($0.352$) & $0.036$ ($0.037$)\\
\hline
\end{tabular}
\end{table}
\subsubsection{Real data}
Next, we perform clustering on real data.
The following three datasets are used:
\begin{itemize}
 \item Accelerometry data: $5$-dimensional data used in
       \citep{hachiya2012importance} for human activity recognition extracted
       from mobile sensing data available from
       \url{http://alkan.mns.kyutech.ac.jp/web/data}.  The number of classes is
       $3$.  In each run of experiment, we use randomly chosen $100$ samples
       from each class.  The total number of samples is $300$.
       
 \item Vowel data: $10$-dimensional data of recorded British English vowel sounds available from
       \url{https://archive.ics.uci.edu/ml/datasets/Connectionist+Bench+(Vowel+Recognition+-+Deterding+Data)}.
       The number of classes is $11$ In each run of experiment, we use randomly chosen $500$.

 \item Sat-image data: $36$-dimensional multi-spectral satellite image
       available from
       \url{https://archive.ics.uci.edu/ml/datasets/Statlog+(Landsat+Satellite)}.
       The number of classes is $6$.  In each run of experiment, we use
       randomly chosen $2000$ samples.
       
 \item Speech data: $50$-dimensional voice data by two French
       speakers~\citep{sugiyama2014information}. The number of classes
       is $2$.  In each run of experiment, we use randomly chosen $200$
       samples from each class. The total number of samples is $400$.
\end{itemize}

For MT-LSLDG, the hyper-parameters are cross-validated using the candidates,
$\sigma\in\{10^{-1},10^{-6/9},10^{-3/9},\dots,10^{12/9},10^{15/9},10^2\}$,
$\lambda\in\{10^{-5},10^{-4},10^{-3},10^{-2},10^{-1},10^0\}$ and
$\gamma\in\{10^{-5},10^{-4},\dots,10^1,10^2\}$,
except that we use relatively small candidate lists
$\sigma\in\{0.5,1,2.5,5,10\}$, $\lambda\in\{0.003,0.01,0.1,1\}$ and
$\gamma\in\{0.1,1,10\}$ for the speech data
since it has large dimensionality and optimization is computationally expensive.
For S- and C-LSLDG, we used the same candidates
of MT-LSLDG for $\sigma$ and $\lambda$. For mean shift clustering, the
Gaussian kernel is employed in KDE, and the bandwidth parameter in the
kernel is selected by 5-fold cross-validation with respect to the log-likelihood
of the density estimate from the same candidates of MT-LSLDG for $\sigma$.

The results are shown in Table~\ref{table:clusterOnReal}. For the
accelerometry data whose dimensionality is only five, S-LSLDGC gives the best
performance and MT-LSLDGC does not improve the performance,
although MT-LSLDGC performs better than C-LSLDGC\@.

On the other hand, for the higher-dimensional dataset, the vowel data, the sat-image data, and the speech data,
the performance of MT-LSLDGC is the best or comparable to the best.
These results indicate that MT-LSLDG is a promising method in mode-seeking clustering especially when the
dimensionality of data is relatively large.

\begin{table}[t]
\caption{Averages (and standard errors) of ARIs on real data.
         In each row, the best and comparable to the best ARI
         in terms of paired t-test with significance level $5\%$
         is emphasized in bold face.
         The number of trials is 100 for the accelerometry data and the sat-image data,
         and is 20 for the speech data.}
 \label{table:clusterOnReal} \centering
 \vspace{3mm}
\begin{tabular}{rcccc}
\hline
dataset ($d$, $n$) & MT-LSLDGC & S-LSLDGC & C-LSLDGC & Mean-shift\\
\hline
\hline
accelerometry ($5$, $300$)
& $0.40$      ($0.01$)
& $\bm{0.53}$ ($0.02$)
& $0.24$      ($0.01$)
& $0.26$      ($0.04$)\\
vowel ($10$, $500$)
& $\bm{0.15}$ ($0.00$)
& $\bm{0.15}$ ($0.00$)
& $\bm{0.15}$ ($0.00$)
& $0.04$ ($0.00$)\\
%$0.1497$ ($0.0013$)
%$0.1488$ ($0.0013$)
%$0.1489$ ($0.0012$)
%$0.0436$ ($0.00033$)
sat-image ($36$, $2000$)
& $\bm{0.48}$ ($0.00$)
& $0.43$ ($0.01$)
& $0.35$ ($0.00$)
& $0.00$ ($0.00$)\\
speech ($50$, $400$)
& $\bm{0.17}$ ($0.02$)
& $0.00$ ($0.00$)
& $0.15$ ($0.01$)
& $0.00$ ($0.00$)\\
\hline
\end{tabular}
\end{table}

%8-----------------------------------------
\section{Conclusion}
\label{sec:conclusion}
We proposed a multi-task log-density gradient estimator in order to improve the existing
estimator in higher-dimensional cases.
Our fundamental idea is to exploit the relatedness inhering in the partial derivatives
through regularized
multi-task learning. As a result, we experimentally confirmed that our
method significantly improves the accuracy of log-density gradient
estimation. Finally, we demonstrated its usefulness of the proposed
log-density gradient estimator in mode-seeking clustering.

Log-density gradient estimation would be useful in a measure for
non-Gaussianity~\citep{huber1985projection} and other further
fundamental statistical topics~\citep{singh1977applications}. In the
future work, we will investigate the performance of our proposed method
in these topics.

%\subsection*{Acknowledgments}
%
%\section*{Appendix}

%References--------------------------------
%\addcontentsline{toc}{section}{Bibliography}
\bibliographystyle{plainnat}
%\bibliography{references}

\end{document}